%% file: main.tex
\documentclass[a4paper]{svproc}

\usepackage{url}
\usepackage{amsmath}
\usepackage{amssymb}
\usepackage{graphicx}
\usepackage{comment}
\usepackage{multirow}
\usepackage[most]{tcolorbox}

\usepackage{tabularray}
\UseTblrLibrary{booktabs}

\begin{document}
\mainmatter              % start of a contribution
\title{OVAL-Grasp: Open-Vocabulary Affordance Localization for Task Oriented Grasping}
\titlerunning{Oval-Grasp}  % abbreviated title (for running head)
\author{Edmond Tong\thanks{Equal contribution}\inst{1} \and Advaith Balaji\footnotemark[1]\inst{1} \and Anthony Opipari\inst{1} \and  \\
Stanley Lewis\inst{1} \and Zhen Zeng\inst{2} \and Odest Chadwicke Jenkins\inst{1}}

\authorrunning{Edmond Tong et al.} % abbreviated author list (for running head)
\tocauthor{Edmond Tong , Advaith Balaji, Anthony Opipari, Stanley Lewis, Zhen Zeng, Odest Chadwicke Jenkins}
\institute{University of Michigan, Ann Arbor, MI, USA\\
\and
J.P. Morgan AI Research}

\maketitle              % typeset the title of the contribution

\input{sec/0_abstract}
\input{sec/1_intro_related}

\input{sec/2_problem}
\input{sec/3_approach}
\input{sec/4_experiments}
\input{sec/6_conclusion}
\bibliographystyle{spmpsci}
\bibliography{bib}
\end{document}

%% file: sec/0_abstract.tex
\begin{abstract}

To manipulate objects in novel, unstructured environments, robots need task-oriented grasps that target object parts based on the given task. Geometry-based methods often struggle with visually defined parts, occlusions, and unseen objects. We introduce \textbf{OVAL-Grasp}, a zero-shot open-vocabulary approach to task-oriented, affordance based grasping that uses large-language models (LLM) and vision-language models (VLM) to allow a robot to grasp objects at the correct part according to a given task. Given an RGB image and a task, OVAL-Grasp identifies parts to grasp or avoid with an LLM, segments them with a VLM, and generates a 2D heatmap of actionable regions on the object. During our evaluations, we found that our method outperformed two task oriented grasping baselines on experiments with 20 household objects with 3 unique tasks for each. OVAL-Grasp successfully identifies and segments the correct object part 95\% of the time and grasps the correct actionable area 78.3\% of the time in real-world experiments with the Fetch mobile manipulator. Additionally, OVAL-Grasp finds correct object parts under partial occlusions, demonstrating a part selection success rate of 80\% in cluttered scenes. We also demonstrate OVAL-Grasp's efficacy in scenarios that rely on visual features for part selection, and show the benefit of a modular design through our ablation experiments. Our project webpage is available at \textcolor{blue}{https://ekjt.github.io/OVAL-Grasp/}.\\


\end{abstract}

%% file: sec/1_intro_related.tex
\vspace{-8mm}

\begin{figure}
    \centering
    \includegraphics[width=1\linewidth]{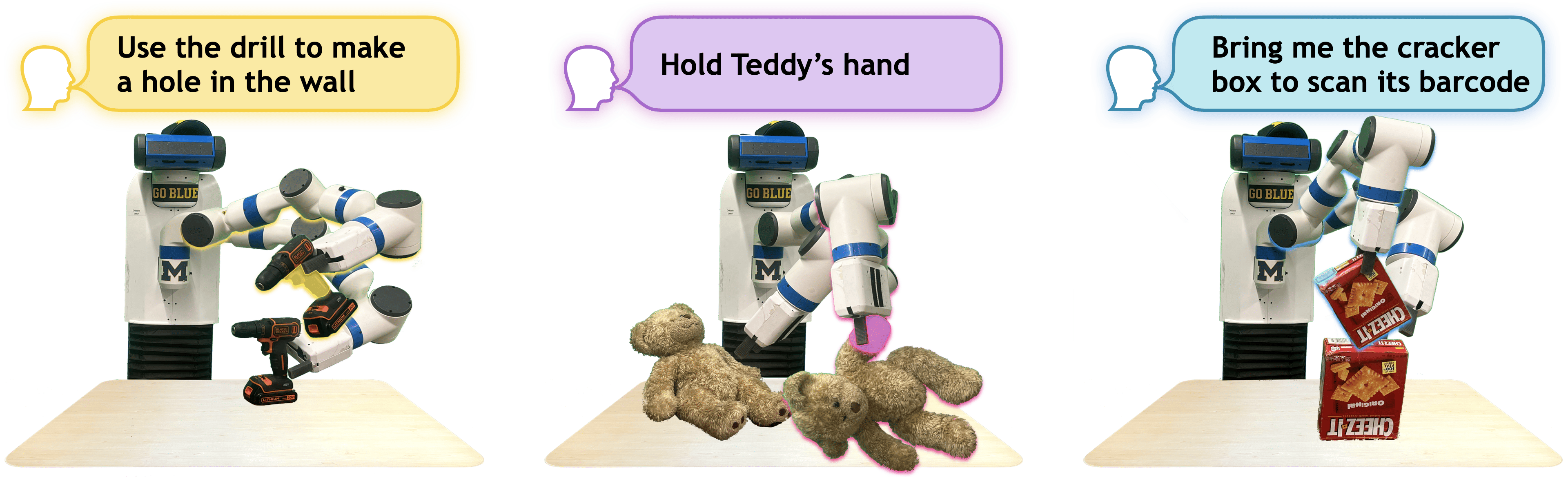}
    \vspace{-25pt}
    \caption{OVAL-Grasp at work on the Fetch mobile manipulator. The robot understands which parts of the object it should grasp and which parts should be avoided to fulfill the given tasks described by language.}
    \label{fig:attention}
\end{figure}

\vspace{-3mm}
\section{Introduction \& Related Work}
\vspace{-3mm}

Robots in unstructured environments must identify objects and understand their \textit{affordances}—the actions they enable. Affordance grounding links actions to object geometry. Most current methods are end-to-end, relying on supervised training with fixed affordances, which limits adaptability~\cite{aff_dl_survey}. Recent work explores one or zero-shot (open vocabulary) affordance localization~\cite{Learningluo,3d_open_aff}. However, they still require some fine-tuning, which can be costly and dataset-dependent. We present our method \textbf{OVAL-Grasp}, a modular and training free approach to task oriented, affordance-based grasping, that enables a robot to grasp an open set of objects at the correct part according to a task described by language. This paper presents our method, along with an evaluation of its performance compared to two state-of-the-art (SOTA) baselines, ShapeGrasp and GraspGPT~\cite{li2024shapegrasp,tang2023graspgpt}. Figure \ref{fig:attention} demonstrates OVAL-Grasp in action.

Large Language Models (LLMs) and Vision Language Models (VLMs) have emerged as powerful open-vocabulary tools \cite{driess2023palme}, excelling in robotics applications like planning and navigation~\cite{song2023llmplanner,shah2022lmnav}. Prior affordance localization approaches used fine-tuned LLMs~\cite{qian2024affordancellm} or text encoders~\cite{3d_open_aff}. In contrast, we explore affordance localization using pre-trained foundation models \textit{without} fine-tuning. Task-Oriented Grasping (TOG) selects grasps based on task input and affordances~\cite{kleeberger2020survey}. Deep learning-based TOG methods~\cite{tang2023task} require supervision, while recent zero-shot approaches~\cite{tang2023graspgpt,tang2025foundationgrasp,li2024shapegrasp} aim to generalize but often still need training. In this work, we present the following contributions: 

\begin{enumerate}
    \item \textbf{OVAL-Grasp}, a method for zero-shot task-oriented grasping by leveraging LLMs and VLMs.
    \item A modular design that enables performance improvements as individual components are enhanced.
    \item A method that achieves superior performance over SOTA methods in real robotic experiments.
    \item Experiments that demonstrate the capability of our method in novel TOG scenarios such as completing tasks that require both visual and geometric cues.
    \vspace{-4mm}
\end{enumerate}

%% file: sec/2_problem.tex
\section{Problem Definition}
\vspace{-3mm}
In order for the robot to perform a successful zero-shot task oriented grasp given an RGB-D image $ I \in \mathbb{R}^{H \times W \times C} $ as input, the robot must generate an end-effector pose $g \in SE(3)$ that lies on a part of an object $x$ that facilitates the execution of a task $t$ described by natural language. We assume that the robot has an egocentric RGB-D camera as well as a manipulator arm with a parallel jaw gripper capable of navigating to the grasp target $g$. We also assume access to a Large Language Model $L$, a vision-language part segmentation model $V$, and a grasp proposal model $G$. TOG methods can be evaluated by measuring their part identification and grasp success rate~\cite{li2024shapegrasp,tang2025foundationgrasp,tang2023graspgpt}. A task oriented grasp is considered successful if the proposed grasp location fulfills the given task based on commonsense reasoning. 

%% file: sec/3_approach.tex
\begin{figure}[t]
    \centering
    \includegraphics[width=1\linewidth]{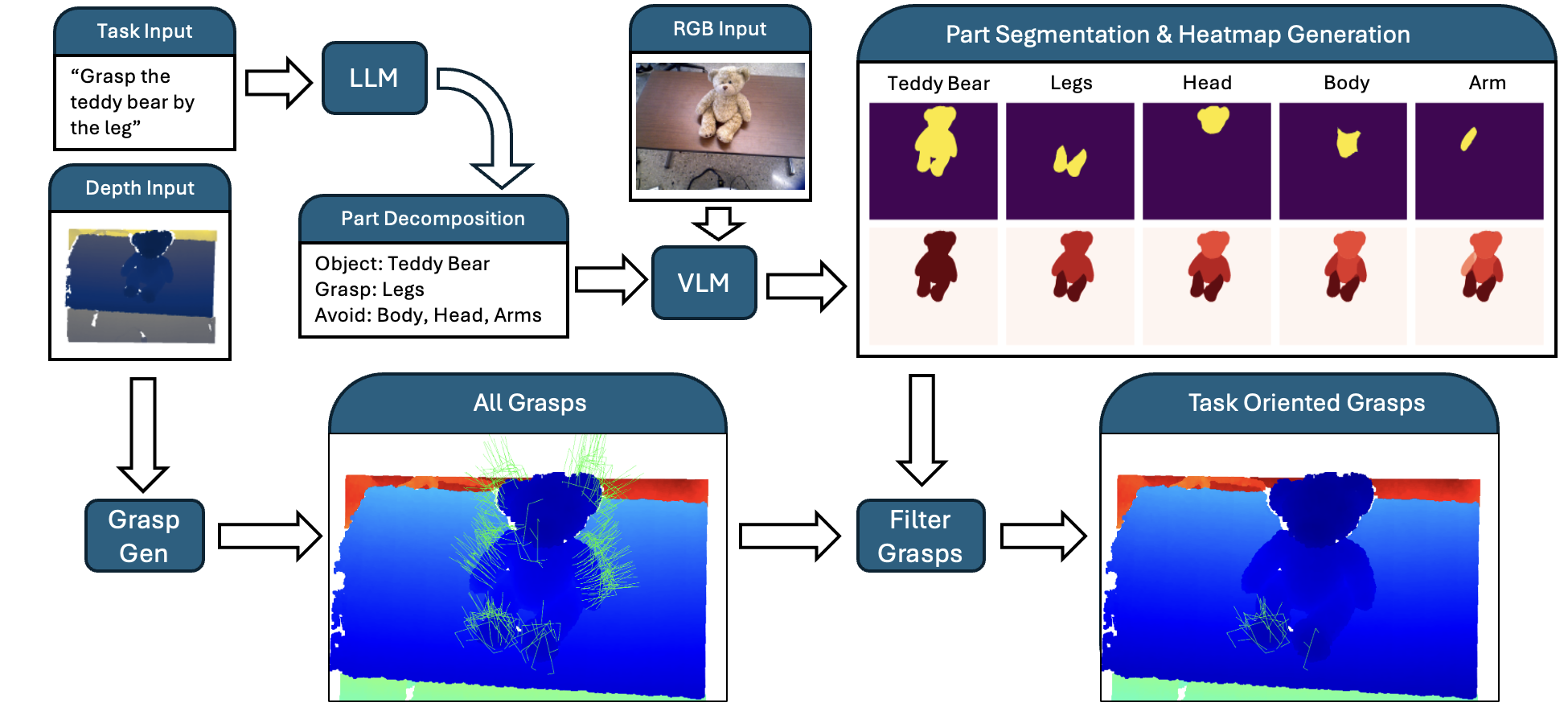}
    \caption{System overview. The robot generates a task-oriented grasps by using an LLM to identify grasp-relevant object parts, a VLM to segment them, and a constructed heatmap to filter grasp candidates to produce a set of grasp that fulfill the given task.}
    \vspace{-5mm}
    \label{fig:system}
\end{figure}

\vspace{-3mm}
\section{OVAL-Grasp: Affordance-Prompting}
\vspace{-3mm}

OVAL-Grasp processes an RGB image and a task description to localize affordances. As shown in Figure \ref{fig:system}, the language model $L$ uses the task $t$ to identify the object $x$, and decompose it into the desirable parts (to grasp), and the undesirable parts (to avoid). The part segmentation model $V$ then segments these parts, which we use to generate a heatmap based on each part's confidence score. The grasp generator $G$ proposes candidate grasps for the entire object, which are subsequently scored using the heatmap.

\vspace{-4mm}
\subsubsection{Part Decomposition:}
We prompt $L$ to identify and decompose the relevant object $x$ into desirable and undesirable parts, where parts that facilitate task $t$ are desirable, and vice-versa. We use the GPT-4o API~\cite{openai2025chatgpt}. 

\vspace{-4mm}
\subsubsection{Part Localization:}
We prompt $V$ with the list of identified parts, producing a binary mask and confidence score for each part. We use PartGLEE~\cite{li2024partglee} for its SOTA open-vocabulary part segmentation capabilities.

\vspace{-4mm}
\subsubsection{Heatmap Generation:}
OVAL-Grasp filters grasp candidates by composing a heatmap from relevant part segments and confidence scores. We generate a heatmap with the same height and width as the input image to score potential grasps. The heatmap $H$ is initialized to zero , and the whole object segmentation is added with a positive value. The desirable part segments are added to the heatmap and undesirable segments are subtracted from the heatmap and are both scaled by their respective confidence scores. The values in the heatmap $H$ were scaled to the range $[0, 255]$ and then smoothed using a Gaussian blur with a $3 \times 3$ kernel to remove any noise from adding segments. 

\vspace{-2mm}
\subsubsection{Grasp Generation and Filtering:}

Based on the previous modules' outputs, OVAL-Grasp scores candidate grasps using the task-specific heatmap, selecting the highest-scoring pose for execution. We used ContactGraspNet~\cite{sundermeyer2021contact} to generate the candidate grasps \(g_c\), where the number of candidates ranges from 0 to N depending on the scene. Each candidate grasp is scored using $H$. We use two scores, the contact score ensures the grasp targets the correct region, while the z-axis score discourages approach angles that could obstruct task-relevant areas. The contact score \( S_c(g_{c_i}) \) is computed by reprojecting the gripper’s contact point \((x_c, y_c)\), provided by ContactGraspNet, into the heatmap \( H \). From the object point cloud \( P \), and the line \( z_g \) created by extending the z-axis of the grasp, we find point \( p \in P \) that is closet to \( z_g \). The z-axis score \( S_z(g_{c_i}) \) is calculated by reprojecting \( p \) to \( H \),  $(x_{p}, y_{p})$. 
\begin{align}
S_c(g_{c_i}) &= H(x_c, y_c) \hspace{0.5cm} S_z(g_{c_i}) = H(x_{p}, y_{p}) \hspace{0.5cm} S(g_{c_i}) = S_c(g_{c_i}) + S_z(g_{c_i})
\label{eqn:score}
\end{align}
The grasps are then sorted by their total score $S(g_{c_i})$ which is calculated as shown in Equation \ref{eqn:score}. The highest score grasp is selected as the grasp target $g$. 

%% file: sec/4_experiments.tex
\vspace{-3mm}
\section{Experiments \& Results}
\begin{figure}[t!]
    \centering
    \includegraphics[width=0.381\linewidth]{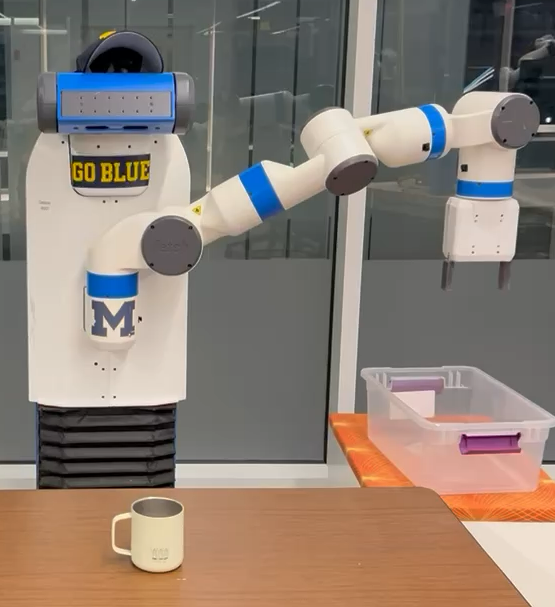}
    \includegraphics[width=0.595\linewidth]{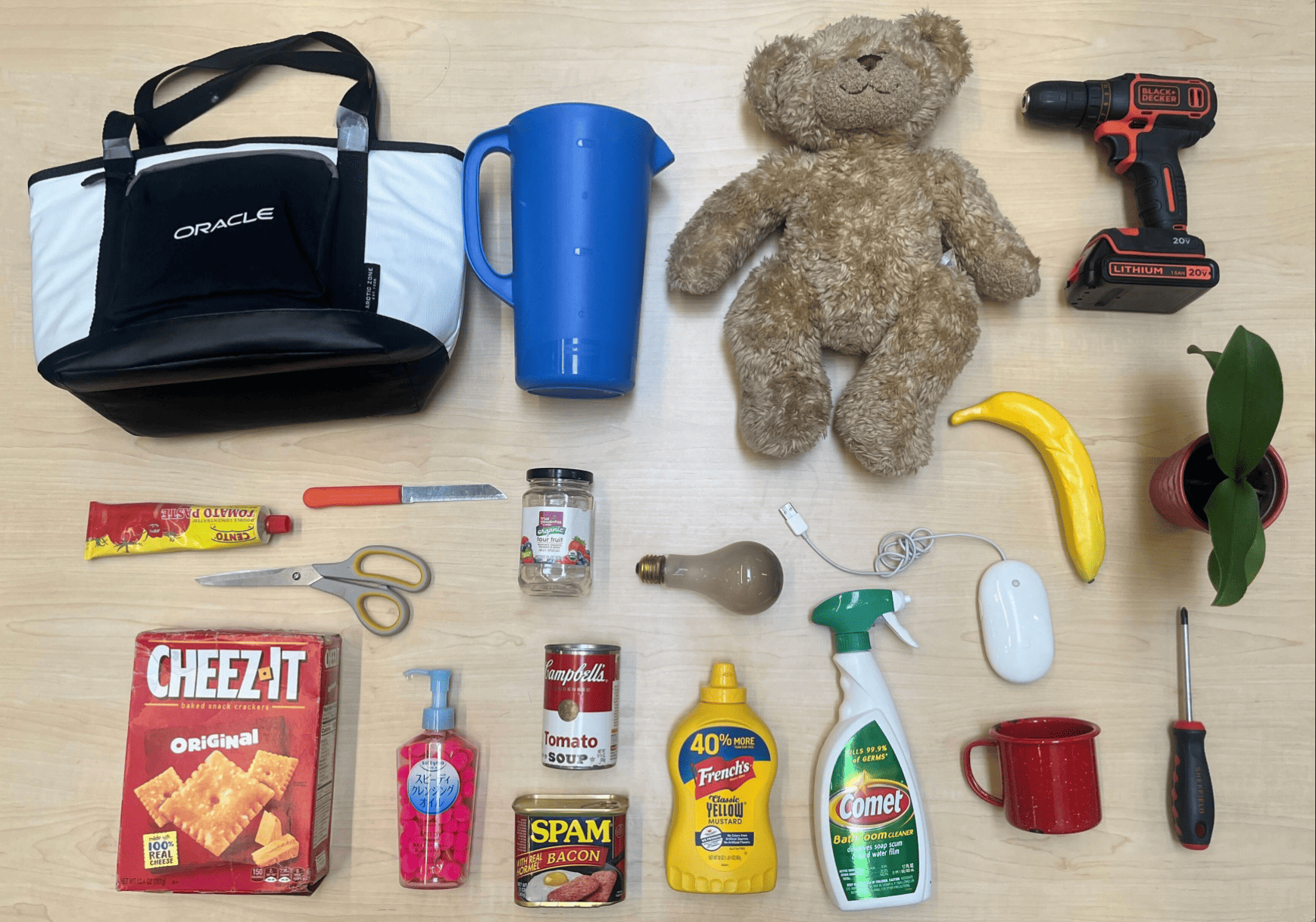}
    \caption{Our experimental setup used the Fetch robot (left) and household and YCB objects (right) to evaluate OVAL-Grasp and baseline methods.}
    \vspace{-5mm}
    \label{fig:setup}
\end{figure}
\vspace{-3mm}
We evaluate OVAL-Grasp in four aspects: (1) part identification and grasp success in table-top settings, (2) robustness in cluttered scenes, (3) vision-based reasoning for task-oriented grasping, and (4) an ablation study isolating the impact of different language and segmentation models.
\vspace{-2mm}
\subsection{Experiment Setup} 
\vspace{-2mm}

We tested our method and a set of baseline task-oriented grasping algorithms using a Fetch and Freight Research Edition robot to grasp objects placed on a table. We compare our model against the baselines GraspGPT \cite{tang2023graspgpt} and ShapeGrasp \cite{li2024shapegrasp}. These models were chosen based on their state-of-the-art performance on task oriented grasping benchmarks. RGB-D images were captured using the onboard head camera. For each method, we evaluate grasp success through multiple trials. We control for object configuration by resetting the objects' locations consistently between each trial and across all the methods tested. It is important to note that both GraspGPT and ShapeGrasp are given the ground truth segmentation mask of the relevant object to allow them to produce a predicted task oriented grasp; our method produces a task oriented grasp directly without needing a ground truth object mask. To evaluate the models, we tested them on 20 objects, each with three tasks requiring the robot to grasp the object in three different locations while positioned in front of them. The object set includes samples from the first and second baseline methods, objects from the YCB dataset \cite{ycb}, and additional objects not demonstrated in either baseline. Figure \ref{fig:setup} shows the experimental setup and the objects tested. Each method was evaluated on two metrics—Part Selection Success and Grasp Success—similar to the metrics in \cite{li2024shapegrasp}. Part Selection Success is the proportion of trials in which the system correctly labels the task-relevant part and (by commonsense visual inspection) generates a mask that covers 100\% of the part’s visible region in the RGB image. Grasp Success is the proportion of trials in which the robot then executes a stable grasp on the identified part and lifts it off the table.

\begin{figure}[t]
    \centering
    \includegraphics[width=1\linewidth]{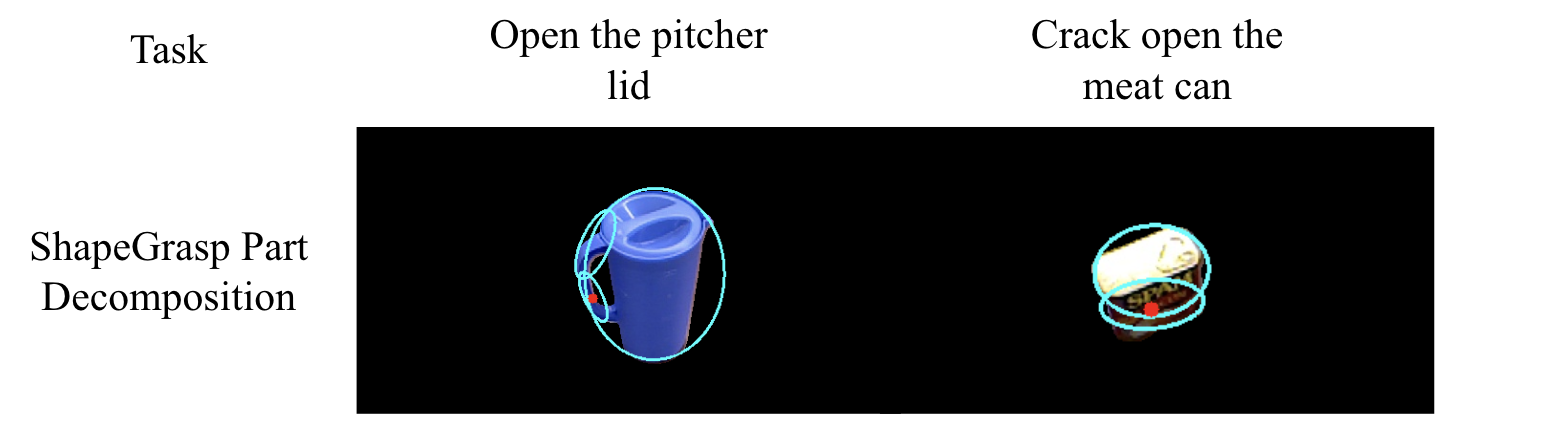}
    \caption{Examples of ShapeGrasp failures. When dealing with more convex geometries and parts that are flush with the object, ShapeGrasp fails to identify the part.}
    \vspace{-4mm}
    \label{fig:shapegrasp_fail}
\end{figure}

\subsection{Grasp Success Experiment}
\vspace{-2mm}
The results of the Grasp Success Experiment are shown in Table \ref{table:comparison}. We find that OVAL-Grasp outperforms the other two baselines, achieving 35\% better performance than GraspGPT and 21.7\% better than ShapeGrasp in part selection, despite not having access to the ground truth object mask. GraspGPT struggles to generalize to unique and novel objects or tasks, frequently scoring all proposed grasps nearly equal. ShapeGrasp performs better but has difficulty decomposing objects with simple convex geometries, contiguous parts or parts flush with the object's body. This is exemplified in Figure \ref{fig:shapegrasp_fail}, where ShapeGrasp is unable to find the pitcher lid, and pull tab of the can. The few part-selection errors for OVAL-Grasp, accounting for 5\% of failed trials, occur when $V$ fails to generate a valid segment or when $L$ hallucinates an incorrect decomposition, whereas the 21.7\% of failed grasps result from $G$ producing only marginal or no grasp candidates.

\begin{table}[t]
\centering
\caption{Task-oriented grasping experiment on 20 objects with 3 tasks per object.}
\label{table:comparison}
\begin{tabular}{p{2.5cm} p{2.4cm} p{3.4cm} p{2.3cm} p{1.1cm}}

\toprule
Method & Segment Source & Part Selection Success & Grasp Success & Time \\
\midrule
GraspGPT \cite{tang2023graspgpt} & Ground truth & 60.0\% & 56.7\% & 128.67s \\
ShapeGrasp \cite{li2024shapegrasp} & Ground truth & 73.3\% & 66.7\% & 22.21s \\
OVAL-Grasp & PartGLEE\cite{li2024partglee} & \textbf{95.0\%} & \textbf{78.3\%} & \textbf{19.86s} \\

\bottomrule
\end{tabular}
\vspace{-5mm}
\end{table}

Our grasping success rate is higher than both baselines. ShapeGrasp’s grasping strategy relies on geometry and heuristics around the part’s centroid, which sometimes leads to grasp failures. GraspGPT’s grasps also have a lower success rate due to poor semantic understanding. In contrast, our method leverages an off-the-shelf grasp generator that proposes a variety of grasps, which are then filtered through an affordance-based heatmap, resulting in more effective overall grasping.

\vspace{-2mm}
\subsection{Runtime Evaluation}
\vspace{-2mm}

To assess runtime efficiency, we measured the average wall-clock time of each method on a system with an RTX 3090 GPU and an Intel i9-11900KF CPU. GraspGPT is an order of magnitude slower than both ShapeGrasp and our method due to repeated LLM queries. In contrast, ShapeGrasp and our approach require only a single query. However, none of the methods currently achieve real-time performance, as LLM inference introduces significant latency. 

\vspace{-3mm}
\subsection{Visual Semantics}
\vspace{-8mm}
\begin{figure}[h]
    \centering
    \includegraphics[width=0.9\linewidth]{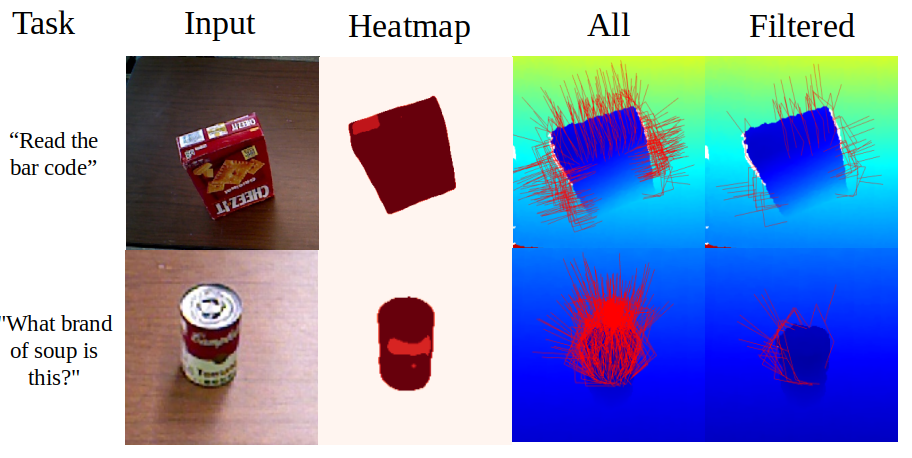}
    \caption{OVAL-Grasp idenitifes object parts not linked to the object's geomtery. Scores are assigned to the barcode and soup can label segments in the heatmap and grasps that obstruct them are filtered out.}
    \label{fig:barcode}
\end{figure}
\vspace{-4mm}
OVAL-Grasp is also able to reason about visual semantics. Some tasks require visual understanding beyond geometry, such as scanning a barcode or reading a label on a can or bottle, a common task in a retail or domestic environment. These parts are defined by visual features (e.g., colors and patterns) and cannot be captured purely by the object's geometry. Unlike other baselines that rely solely on geometry, our system uses visual RGB input to segment parts, enabling more context-aware grasping. Figure \ref{fig:barcode} illustrates an example of this scenario, where our approach is able to correctly identify and avoid grasping the barcode and can label, both of which do not exist as part of the geometry.

\vspace{-4mm}
\subsection{Cluttered Scenes}
\begin{figure}[t!]
    \centering
    \includegraphics[width=0.9\linewidth]{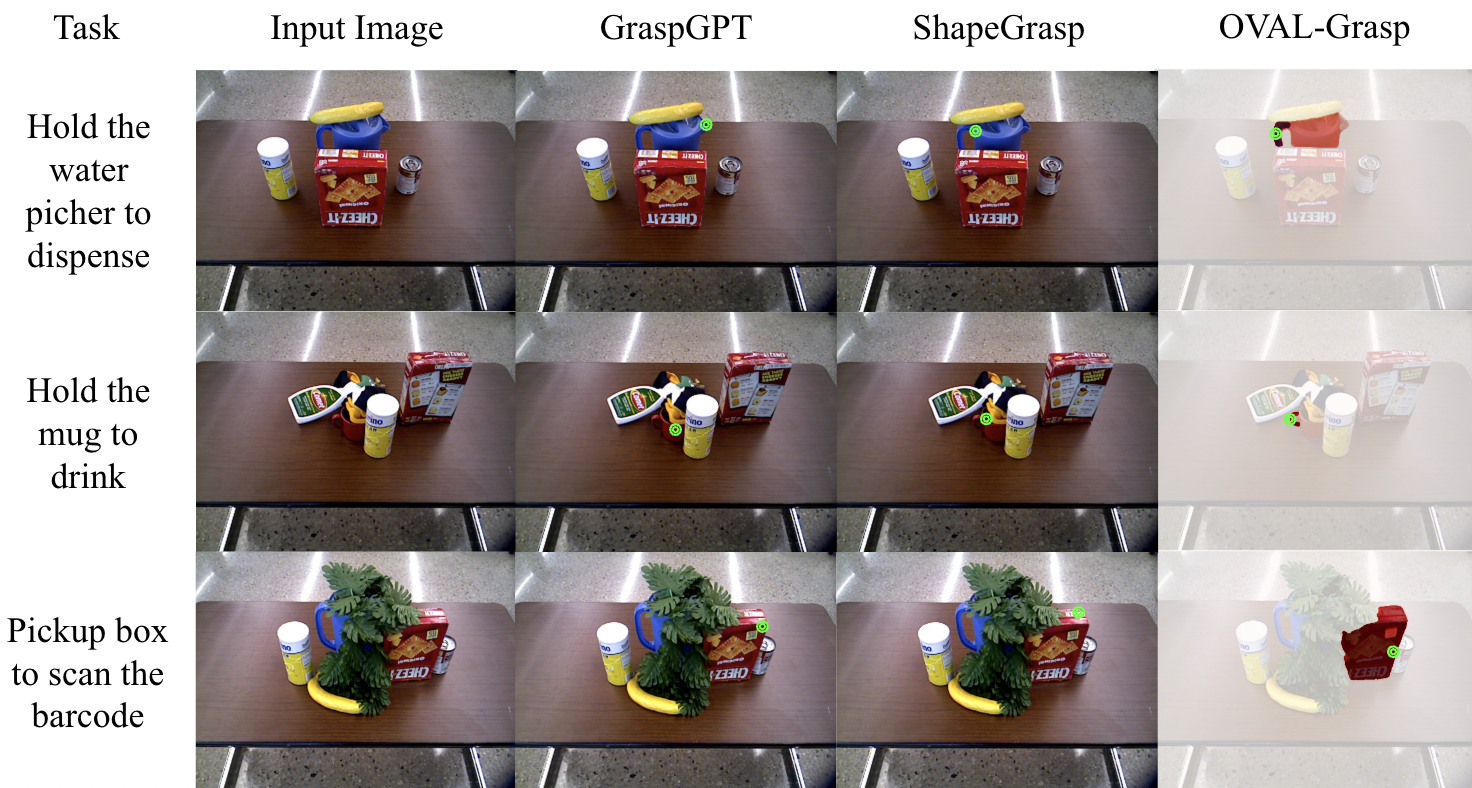}
    \caption{Examples comparing OVAL-Grasp and baselines on part identification and grasp proposal in clutter. Proposed grasp point is highlighted in green. OVAL-Grasp successfully identifies the grasp point even with partially occluded parts.}
    \label{fig:clutter}
\end{figure}

\vspace{-2mm}
Next, we set out to test each of the considered algorithms in cluttered scenarios that are common in everyday human environments. To do this, we evaluated our method and baseline approaches in cluttered scenes containing multiple objects with varying degrees of occlusion and overlap. Each method was tested on a similar set of tasks and objects as in the main experiment to ensure a fair comparison. While cluttered scenes fall outside the original scope of the other two methods, we included them to assess how robust our model would be in real-world settings. For this evaluation, we focused only on Part Selection Success, not on the success of the final grasp execution. The results are reported in Table \ref{table:clutter} and are visualized in Figure \ref{fig:clutter}.

\begin{table}[t]
\centering
\caption{Task-oriented grasping experiment on 15 cluttered scenes}
\label{table:clutter}
\begin{tabular}{p{2.5cm} p{2.4cm} p{3.4cm} }
\toprule
Method & Segment Source & Part Selection Success   \\
\midrule
GraspGPT \cite{tang2023graspgpt} & Ground truth & 26.7\%  \\
ShapeGrasp \cite{li2024shapegrasp} & Ground truth & 46.7\%  \\
OVAL-Grasp & PartGLEE\cite{li2024partglee} & \textbf{80.0\%} \\

\bottomrule
\end{tabular}
\vspace{-6mm}
\end{table}

OVAL-Grasp performed robustly in cluttered scenes with occluded objects, because it does not rely solely on geometric reasoning. In contrast, geometry-dependent methods like GraspGPT and ShapeGrasp struggled when the target object was partially hidden or fragmented. These methods rely on clear object visibility and intact shape priors: GraspGPT, trained on complete point clouds, failed to generalize to fragmented inputs, while ShapeGrasp required canonical parts to be visible for accurate part identification. Both methods were prone to misidentifying object parts in clutter or when disjointed configurations made a single object appear as multiple parts. OVAL-Grasp, however, remained effective under occlusion and appearance shifts, successfully identifying task-relevant grasps even when parts were only partially visible or distinguishable by visual features alone.

\vspace{-2mm}
\subsection{Ablation on Modular Design}
\vspace{-2mm}
To evaluate the contribution of each component, we conducted an ablation study by replacing the part segmentation and language models. Using a subset of objects from the main experiments, we assessed performance based on part selection results.

\vspace{-4mm}
\begin{table}[h]
\centering
\caption{Ablation study demonstrating OVAL-Grasp's performance with different part segmentation models and LLMs}
\label{table:Ablation}
\begin{tabular}{p{2.3cm} p{2.8cm} p{3.1cm} p{3.4cm}}
\toprule
 Method & Segment Source & Language Model & Part Selection Success \\
\midrule
\multirow{4}{*}{OVAL-Grasp} & PartGLEE & Deepseek-R1 (7B) & 33.3\% \\
                            & PartGLEE & GPT-3.5 Turbo & 58.3\% \\
\cmidrule(lr){2-4}     
                            & VLPart & GPT-4o & 78.3\% \\
                            & PartGLEE & GPT-4o & \textbf{95.0\%} \\

\bottomrule
\end{tabular}
\vspace{-6mm}
\end{table}

Table~\ref{table:Ablation} contains the results of our ablation experiments. We tested 3 language models—GPT-4o~\cite{openai2025chatgpt}, GPT-3.5 Turbo~\cite{openai_gpt3.5_turbo}, and DeepSeek-R1~\cite{deepseekai2025deepseekr1incentivizingreasoningcapability} with the best-performing part segmentation model, PartGLEE~\cite{li2024partglee}. We also tested 2 segmentation models, PartGLEE and VLPart~\cite{vlpart}, with the best performing language model GPT-4o. On the language model side, GPT-3.5 Turbo struggled with part-level reasoning, defaulting to object-level descriptions and showing higher rates of hallucination and prompt sensitivity. DeepSeek-R1 performed worst, likely due to its smaller model size (7B) and incorrect reasoning. These issues align with evidence that chain-of-thought reasoning can impair visual understanding when language struggles to capture spatial or part-level concepts~\cite{mindstepbystep}. On the segmentation side, VLPart underperformed compared to PartGLEE due to lower part detection accuracy. PartGLEE's hierarchical modeling of object parts contributed to more reliable affordance localization. These results suggest that continued improvements in LLM reasoning and VLM segmentation will directly enhance system performance.

\begin{figure}[t]
    \centering
    \includegraphics[width=.7\linewidth]{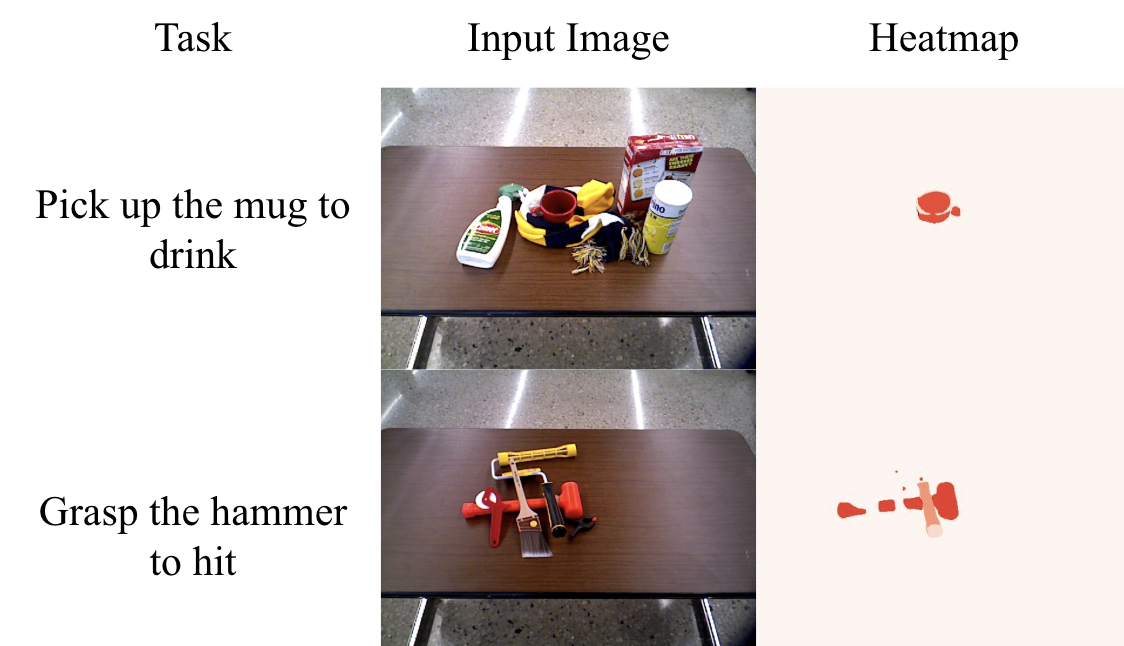}
    \caption{Examples of part identification failures in clutter}
    \vspace{-5mm}
    \label{fig:failure_cases_clutter}
\end{figure}

Despite its strong performance, OVAL-Grasp remains subject to hallucinations and other limitations of large pretrained models. Most failures stem from incorrect part decomposition by the LLM, either hallucinating non-existent parts or misclassifying graspable and ungraspable regions. Figure~\ref{fig:failure_cases_clutter} shows two failure cases in cluttered scenes: one over-segments a mug due to high clutter, and the other incorrectly segments multiple handles from overlapping objects. In clutter, the segmentation model becomes the main source of error. While LLM hallucinations still occur, occlusion often causes the segmentation model to misidentify or over-segment parts, even when given the correct part name. As our ablation studies show, these failures should decrease as individual components improve.

%% file: sec/6_conclusion.tex
\vspace{-2mm}
\section{Conclusion}
\vspace{-3mm}

In this paper, we introduced OVAL-Grasp, a zero-shot task-oriented grasping framework that uses large language and vision-language models for open-vocabulary affordance localization. OVAL-Grasp outperforms GraspGPT and ShapeGrasp in part selection and grasp success, especially in cluttered scenes and visual part identification scenarios. By leveraging VLMs, it enables context-aware grasping beyond geometric cues. Its modular design ensures scalability as foundation models improve, advancing zero-shot grasping in robotics. The main limitations of our method stem from OVAL-Grasp's open-loop reasoning method. It lacks feedback mechanisms to detect and recover from failed or incorrect grasps, preventing re-identification or regrasping with new information. The system also supports only single-step grasps and cannot plan multi-stage manipulations or recovery actions. Its reliance on pre-trained models reduces robustness, especially for niche or underrepresented object parts. Future work will integrate closed-loop feedback—using tactile, force, or visual signals to detect failures and active strategies like object reorientation, camera repositioning, or probing to improve part segmentation and affordance localization in unstructured environments.

\vspace{-2mm}
\subsection{Acknowledgement}
\vspace{-2mm}
This work was supported in part by a Qualcomm Innovation Fellowship, Amazon, and
Ford Motor Company. We extend our gratitude to our sponsors.

\vspace{-3mm}